# Localized Definitions and Distributed Reasoning:

# A Proof-of-Concept Mechanistic Interpretability Study via Activation Patching

Nooshin Bahador

Krembil Research Institute, University Health Network, Toronto, Canada.

- The fine-tuned GPT-2 models and related resources are openly accessible on Hugging Face.
- The source code and implementation details are available in the GitHub repository.

**Abstract**

This study investigates the localization of knowledge representation in fine-tuned GPT-2 models using Causal Layer Attribution via Activation Patching (CLAP), a method that identifies critical neural layers responsible for correct answer generation. The model was fine-tuned on 9,958 PubMed abstracts (epilepsy: 20,595 mentions, EEG: 11,674 mentions, seizure: 13,921 mentions) using two configurations with validation loss monitoring for early stopping. CLAP involved (1) caching clean (correct answer) and corrupted (incorrect answer) activations, (2) computing logit difference to quantify model preference, and (3) patching corrupted activations with clean ones to assess recovery. Results revealed three findings: First, patching the first feedforward layer recovered 56% of correct preference, demonstrating that associative knowledge is distributed across multiple layers. Second, patching the final output layer completely restored accuracy (100% recovery), indicating that definitional knowledge is localised. The stronger clean logit difference for definitional questions further supports this localized representation. Third, minimal recovery from convolutional layer patching (13.6%) suggests low-level features contribute marginally to high-level reasoning. Statistical analysis confirmed significant layer-specific effects ($p<0.01$). These findings demonstrate that factual knowledge is more localized and associative knowledge depends on distributed representations. We also showed that editing efficacy depends on task type. Our findings not only reconcile conflicting observations about localization in model editing but also emphasize on using task-adaptive techniques for reliable, interpretable updates.

## 1. Introduction

Model editing refers to techniques that modify a neural network's behavior post-training—such as updating factual knowledge or correcting errors—without full retraining. A common assumption is that successful edits require precise localization, i.e., targeting specific layers where knowledge is concentrated. However, experimental results demonstrate that editing success and localization are actually uncorrelated, challenging the hypothesis that precise mechanistic targeting is necessary for reliable updates (Hase et al., 2023). This aligns with the broader principle of distributed representations in neural systems, observed in both biological and artificial networks.

Neuroscientific research supports this idea, revealing phenomena like mixed selectivity (where neurons respond to multiple stimulus dimensions) and sparse selectivity (where specialized regions function within larger distributed networks). Similarly, vision neuroscience studies show that object recognition is not purely hierarchical or category-based but instead relies on behaviorally relevant, distributed dimensions (Contier et al., 2024). Together, these findings challenge the traditional view that cognitive—or computational—functions must be localized, whether in biological brains or artificial neural networks.

Another human neuroscience research supports a hybrid model of knowledge representation. Explicit recall (conscious retrieval) relies on domain-specific neural regions, while implicit recognition (automatic familiarity detection) engages distributed domain-general networks. Thus, the distinction between these processes depends on task demands (Shehzad et al., 2022).

Like human visual knowledge representation, Question Answering (QA) tasks can be also categorized along a spectrum of complexity, where different question types impose varying computational demands. These range from simple Factual Recall (Single-Hop QA), requiring no reasoning—only memory or lookup—to more complex Associative Reasoning (Bridge Multi-Hop QA), which necessitates linking multiple facts (Balesni et al., 2024).

Based on the above claims, the human brain processes knowledge in both localized and distributed ways, depending on whether it involves explicit recall or implicit recognition. Similarly, question answering exhibits comparable patterns, distinguishing between factual recall and associative reasoning. To explore this further, researchers have identified localized internal computations during factual recall (Sharma et al., 2024). This localized activity contrasts with the more distributed neural networks involved in associative reasoning, which recruit broader areas to integrate and synthesize information. In fact, researchers have shown that model's learned concept representations are not localized but distributed across multiple internal components. This means no single part holds complete knowledge; instead, concepts emerge from interconnected activations (Chang et al., 2023).

Recent advances in model editing have challenged the assumption that precise localization is necessary for modifying neural network behavior, revealing instead that successful edits can occur even with distributed interventions (Hase et al., 2023). Yet, the relationship between knowledge representation, task complexity, and effective editing remains poorly understood. While neuroscience suggests a hybrid model of cognition—where explicit recall relies on localized regions and implicit recognition engages distributed networks (Shehzad et al., 2022)—it is unclear whether AI systems exhibit similar task-dependent encoding. This work bridges these gaps by introducing a task-aware framework for model editing, grounded in analysis of where and how knowledge is stored in language models. Using Causal Layer Attribution via Activation Patching (CLAP), we show a hierarchical knowledge organization: factual recall is localized while associative reasoning involves distributed intermediate representations, with implications for mechanistic interpretability in domain-specific language models and targeted debugging/editing.

## 2. Method

The technique used in this work localize where in the model the "knowledge" or "reasoning" about the correct answer resides. By observing which patched layers restore the model's preference for the correct answer, we can infer which parts of the network are responsible for that behavior.

### 2.1. Data Preparation

The GPT-2 model was trained on a curated subset of 9,958 PubMed abstracts, with domain-specific terms occurring at high frequencies: epilepsy (20,595 mentions), EEG (11,674 mentions), and seizure (13,921 mentions). The dataset was systematically extracted and preprocessed using the [pubmed_meta_analyzer tool](), an automated Python-based pipeline for PubMed metadata retrieval, deduplication, and keyword analysis. The dataset was structured in a tabular format with the following columns: PMID, Title, Abstract, Journal, Keywords, and URL. The GPT-2 model was fine-tuned on textual abstracts extracted from the 'Abstract' column of this dataset. Rows with missing values (NaN) were dropped to ensure data quality, and the remaining abstracts were converted into Hugging Face Dataset objects with a single "text" field for streamlined processing.

### 2.2. Model Fine-Tuning

Table 1 formalizes the GPT-2 fine-tuning (Configuration I) pipeline, where text data D is tokenized into sequences $\mathbb{Z}^n$, optimized via gradient descent $(\theta_{t+1} = \theta_t - \eta \nabla_\theta \mathcal{L}(\theta_t))$ with hyperparameters $\{\eta = 5 \times 10^{-5}, \lambda = 0.01, B = 8\}$, and evaluated using validation loss $\mathcal{L}_{val}$, while memory constraints are managed through chunking (L=512) and subprocess isolation.

**Table 1:** GPT-2 Fine-Tuning Configuration I

| Component | Mathematical Representation | Description |
| --- | --- | --- |
| Base Model | $f_\theta : \mathbb{Z}^n \to \mathbb{R}^{n \times \|V\|}$ | Pretrained GPT-2 (causal LM), where V is the vocabulary. |
| Tokenization | $T : text \to \mathbb{Z}^n$ | Tokenizer mapping text to token IDs |
| Chunking | $chunk(x, L = 512) = \{x_{i:i+L}\}_{i=0}^{\lceil n/L \rceil}$ | Splits text into fixed-length sequences (L=512 tokens) |
| Padding | $pad(x) = x \oplus \{pad\_id\}^{L - \|x\|}$ | Appends pad-token-id to reach length L |
| Data Split | $D_{train}, D_{val} \sim train-test-split(D, \tau = 0.2, seed = 42)$ | 80/20 stratified split |
| Batch Processing | $B_{train} = 8, B_{val} = 2$ | Batch sizes for training/validation |
| Training Loop | epochs=30, early-stop(k=3) | Stops if validation loss doesn't improve for 3 epochs. |
| Learning Rate | $\eta = 5 \times 10^{-5}$ | AdamW optimizer with fixed learning rate. |
| Weight Decay | $\lambda = 0.01$ | L2 regularization strength |
| Warmup Steps | $\eta_t = \eta \cdot min(t/500, 1)$ | Linear warmup over 500 steps |
| Optimization | $\theta_{t+1} = \theta_t - \eta \nabla_\theta \mathcal{L}(\theta_t)$ | AdamW + FP16 + gradient accumulation (4 steps). |
| Early Stopping | Stop if $\mathcal{L}_{val}$ doesn't improve for k = 3 epochs | Monitors validation loss for convergence. |
| Evaluation Metric | $\mathcal{L}_{val} = -\frac{1}{\|D_{val}\|} \sum_{x \in D_{val}} \log P(x)$ | Cross-entropy loss on validation set. |
| Memory Management | $Clear-memory(): GPU\ cache \leftarrow \emptyset$ | Regular GPU cache clearing and garbage collection. |
| Parallelism | Num-proc=4 | Multi-processing for dataset tokenization. |

Table 2 formalizes the GPT-2 fine-tuning (Configuration II) pipeline, where text data D is tokenized into sequences $\mathbb{Z}^n$, optimized via gradient descent $(\theta_{t+1} = \theta_t - \eta \nabla_\theta \mathcal{L}(\theta_t))$ with hyperparameters $\{\eta = 2 \times 10^{-5}, \lambda = 0.1, B = 8\}$, and evaluated using validation loss $\mathcal{L}_{val}$, while memory constraints are managed through chunking (L=512) and subprocess isolation.

**Table 2:** GPT-2 Fine-Tuning Configuration II

| Component | Mathematical Representation | Description |
| --- | --- | --- |
| Base Model | $f_\theta : \mathbb{Z}^n \rightarrow \mathbb{R}^{n \times \|V\|}$ | Pretrained GPT-2 (causal LM), where V is the vocabulary. |
| Tokenization | $T : text \rightarrow \mathbb{Z}^n$ | Tokenizer mapping text to token IDs |
| Chunking | $chunk(x, L = 512) = \{x_{i:i+L}\}_{i=0}^{\lceil n/L \rceil}$ | Splits text into fixed-length sequences (L=512 tokens) |
| Padding | $pad(x) = x \oplus \{pad\_id\}^{L - \|x\|}$ | Appends pad-token-id to reach length L |
| Data Split | $D_{train}, D_{val} \sim train - test - split(D, \tau = 0.2, seed = 42)$ | 80/20 stratified split |
| Batch Processing | $B_{train} = 8, B_{val} = 2$ | Batch sizes for training/validation |
| Training Loop | epochs=10, early-stop(k=5) | Stops if validation loss doesn't improve for 5 epochs. |
| Learning Rate | $\eta = 2 \times 10^{-5}$ | AdamW optimizer with fixed learning rate. |
| Weight Decay | $\lambda = 0.1$ | L2 regularization strength |
| Warmup Steps | $\eta_t = \eta \cdot min(t/500, 1)$ | Linear warmup over 500 steps |
| Optimization | $\theta_{t+1} = \theta_t - \eta \nabla_\theta \mathcal{L}(\theta_t)$ | AdamW (no FP16) + grad accum (4 steps) |
| Early Stopping | Stop if $\mathcal{L}_{val}$ doesn't improve for k = 5 epochs | Monitors validation loss for convergence. |
| Evaluation Metric | $\mathcal{L}_{val} = -\dfrac{1}{\|D_{val}\|} \sum_{x \in D_{val}} \log P(x)$ | Cross-entropy loss on validation set. |
| Memory Management | $Clear-memory(): GPU\ cache \leftarrow \emptyset$ | Regular GPU cache clearing and garbage collection. |
| Parallelism | Num-proc=4 | Multi-processing for dataset tokenization. |

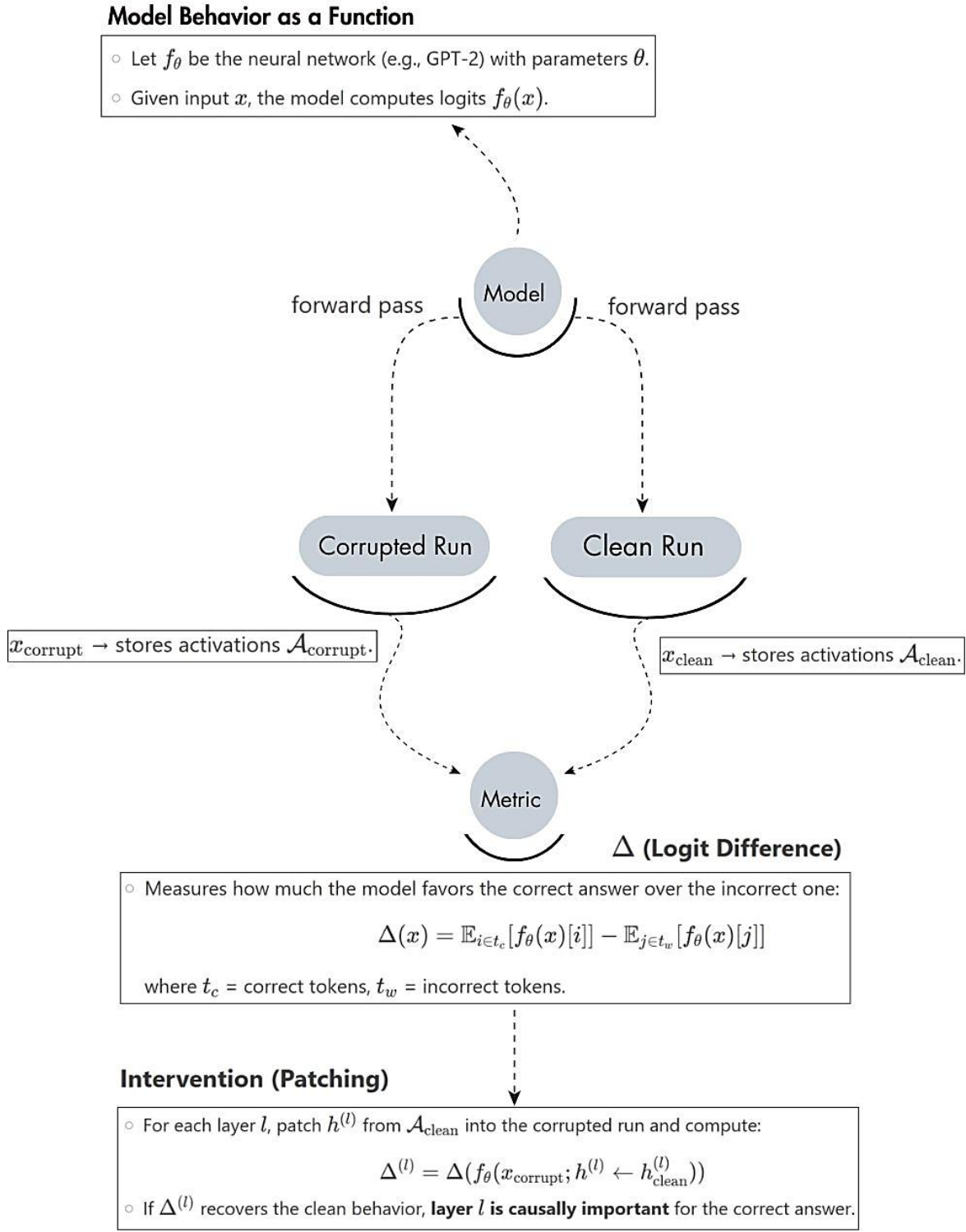

**Figure 1:** Flowchart of the Causal Layer Attribution via Activation Patching (CLAP) algorithm. The process involves: (1) running clean and corrupted inputs through the model to cache activations, (2) computing the logit difference ($\Delta$) to quantify model preference, and (3) patching corrupted activations with clean ones to identify which layers causally restore the correct behavior. Layers that significantly recover $\Delta$ when patched are deemed critical for the model's decision.

## 2.3. Causal Layer Attribution via Activation Patching (CLAP)

CLAP identifies which neural network layers causally influence a model's preference for correct vs. incorrect answers by surgically replacing corrupted activations with clean ones and measuring the recovery in logit difference. As illustrated in the Figure 1, the algorithm tests how patching activations from a "clean" run (correct answer) into a "corrupted" run (incorrect answer) restores model performance and the logit difference metric quantifies the effect.

Step 1) Model and Tokenizer Setup

Let:

- $Tokenizer = T: text \to \mathbb{Z}^n$ (maps text to token IDs)
- $Model = f_\theta: \mathbb{Z}^n \to \mathbb{R}^{n \times |V|}$ (maps token IDs to logits, where $|V|$ is vocabulary size)

Step 2) Input Construction

For question $q$ correct answer $a_c$, corrupted answer $a_w$:

- $x_{clean} = T(q \oplus a_c)$ (clean input sequence)
- $x_{corrupt} = T(q \oplus a_w)$ (corrupted input sequence)

Step 3) Forward Pass with Activation Caching

For any input $x$, the model computes:

- $h(0) = E(x)$ (token embeddings)
- $h^{(l)} = TransformerBlock^{(l)}(h^{(l-1)})$   for   $l = 1, \dots, L$
- $logits = W_{head} h^{(L)}$

where $L = $ number of layers.

Activation Cache stores:

$$A = \{h^{(l)}\}_{l=1}^{L} \cup \{attention\_weights^{(l)}\}_{l=1}^{L}$$

Step 4) Logit Difference Metric

For target token sets $t_c$ (correct) and $t_w$ (incorrect):

$$logit_{diff} = \frac{1}{|t_c|} \sum_{i \in t_c} logits[-1, i] - \frac{1}{|t_w|} \sum_{i \in t_w} logits[-1, i]$$

- mean correct logit : $\left\{\frac{1}{|t_c|}\sum_{i \in t_c} logits[-1, i]\right\}$
- mean incorrect logit : $\left\{\frac{1}{|t_w|}\sum_{i \in t_w} logits[-1, i]\right\}$

This measures how much the model favors the correct phrase over the incorrect one.

Step 5) Activation Patching

For layer $l$:

1. Run corrupted input $x_{corrupt}$ but replace $A^{(l)}$ with the clean version
2. Compute the resulting logit difference

Mathematically, we compute:

- $logit\_diff^{(l)} = metric\left(f_\theta\left(x_{\text{corrupt}}; A^{(l)} \leftarrow A^{(l)}_{clean}\right)\right)$

Where the notation means we substitute the $l_{th}$ layer's activations from the clean run during the corrupted input processing.

## 3. Results

We fine-tuned GPT-2 on a dataset of 9,958 PubMed abstracts, evaluating two distinct training configurations. The training performance was analyzed for each configuration to assess loss dynamics.

For Configuration I, the initial parallel decline in losses (Figure 2) indicates effective learning, but the post-epoch-7 divergence—where training loss falls more sharply than validation loss—suggests the model starts memorizing training specifics rather than generalizing.

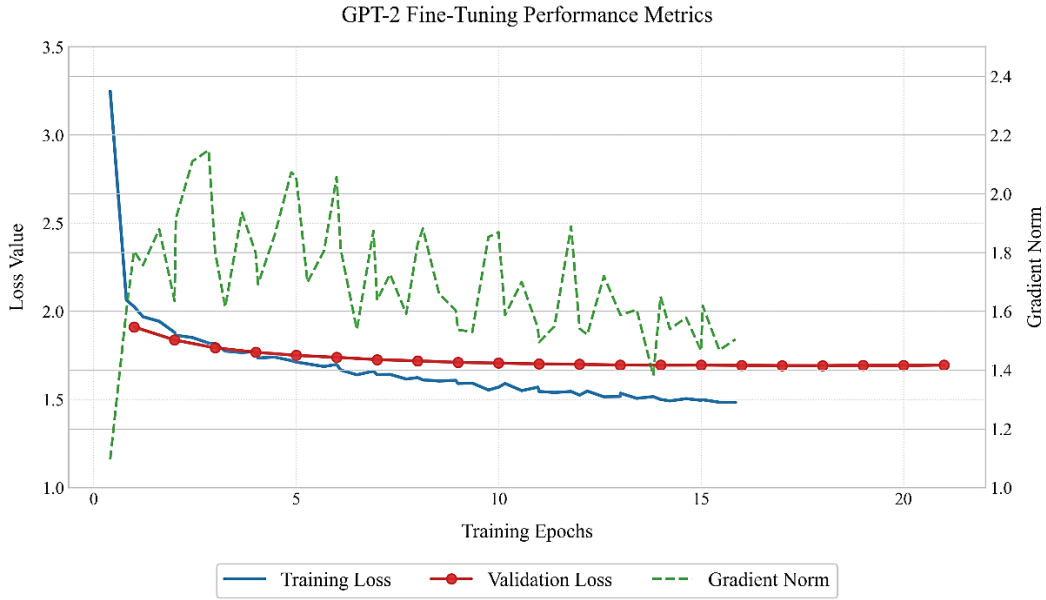

**Figure 2.** Training and validation loss curves alongside gradient norms for fine-tuned GPT-2 (Configuration I)

For Configuration II, the training metrics visualization (Figure 3) reveals that both training and validation losses decrease monotonically while maintaining a consistent gap, indicating effective learning without overfitting, as supported by stable gradient norms throughout the optimization process.

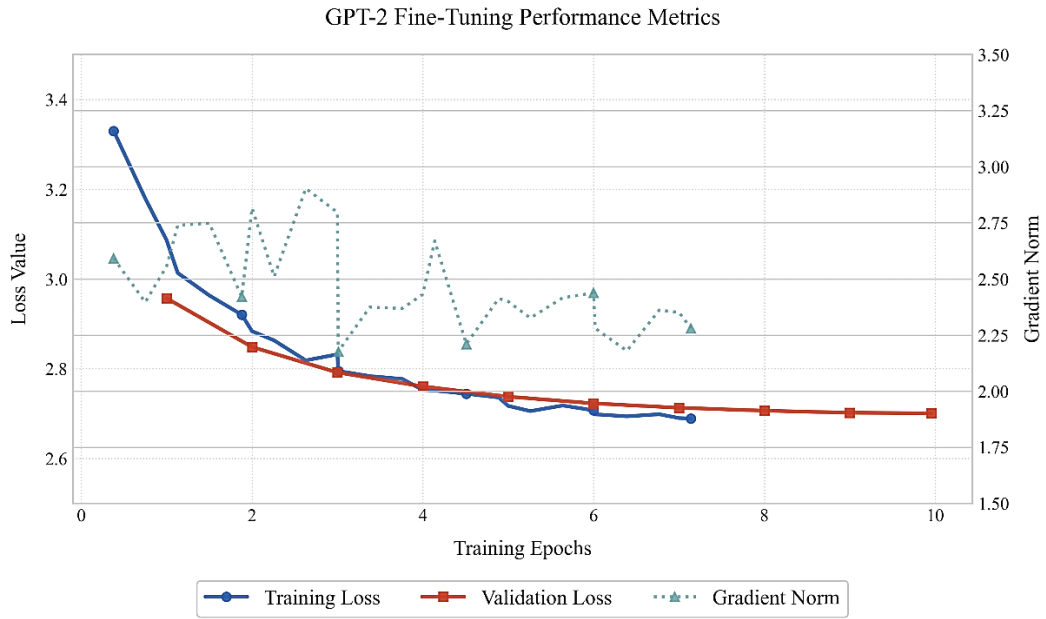

**Figure 3.** Training dynamics of GPT-2 fine-tuning (Configuration II) showing training loss (blue), validation loss (red), and gradient norm (teal) across epochs. The parallel decline in training and validation loss indicates successful model convergence, while stable gradient norms suggest well-behaved optimization.

For activation patching experiments, we used a fine-tuned GPT-2 model (Configuration I, optimized for lower train and validation losses) to investigate the representational encoding of epilepsy-related knowledge in the network. We conducted multiple activation patching experiments on fine-tuned GPT-2, with representative examples shown below.

Table 3 presents activation patching results on GPT-2, analyzing how knowledge of interictal epileptiform discharges is representationally encoded. The table demonstrates that replacing corrupted activations ($\Delta = -0.34$) with clean ones ($\Delta = 0.76$) in the first feedforward layer recovered 56% of correct answer preference ($\Delta = 0.43$).

**Table 3:** Activation Patching Analysis of Medical Fact Retrieval in GPT-2

| Component | Description | Mathematical Syntax |
|---|---|---|
| Question (q) | "What is the presence of interictal epileptiform discharges associated with? | - |
| Correct Answer ($a_a$) | "Increased seizure risk in both focal and generalized epilepsy." | $t_a = tokenizer(a_a)$ |
| Corrupted Answer ($a_x$) | "Cognitive dysfunction risk in both focal and generalized epilepsy." | $t_x = tokenizer(a_x)$ |
| Input Construction | Clean vs. corrupted inputs | $x_{clean} = T(q \oplus a_a), x_{corrupt} = T(q \oplus a_x)$ |
| Target Layer | First feedforward layer (mlp.c_fc) | $h^{(l)}$ where $l = mlp.c\_fc$ |
| Logit Difference Metric | Measures model's preference for correct answer | $\Delta(x) = \mathbb{E}[f_{\theta(x)}[-1, t_a]] - \mathbb{E}[f_{\theta(x)}[-1, t_x]]$ |
| Clean Run | Model's output on correct answer | $\Delta(x_{clean}) = 0.76$ |
| Corrupted Run | Model's output on corrupted answer | $\Delta(x_{corrupt}) = -0.34$ |
| Activation Patching | Replace corrupted activations with clean ones | $f_\theta(x_{corrupt}; h^{(l)} \leftarrow h_{clean}^{(l)})$ |
| Patched Result | Partial recovery of correct preference after patching | $\Delta_{patched} = 0.43$ (56% $recovery$: $(0.43 - (-0.34))/(0.76 - (-0.34)) = 0.56$) |

As shown in Table 4, activation patching of GPT-2's final output layer ($W_{out} \in \mathbb{R}^{\{768 \times 50257\}}$) completely restored the model's ability to correctly define IEDs - patching this single layer increased the logit difference from Δ=0.9691 (corrupted) to Δ=2.1928 (clean), achieving 100% recovery. This demonstrates that: (1) the final projection layer alone mediated the model's definitional knowledge in this case, as evidenced by the perfect Δ restoration ($\partial\Delta/\partial W_{out} \approx 1$); (2) the higher clean Δ (2.19 vs previous 0.76) indicates stronger baseline confidence in medical facts than associations; and (3) unlike the partial recovery (56%) seen when patching intermediate layers in the epilepsy risk experiment, this full recovery suggests compressed representation of definitional knowledge in the output weights.

Table 4: Activation Patching Results for Medical Abbreviation Knowledge in GPT-2

| Component | Description | Mathematical Syntax |
|---|---|---|
| Question (q) | "What is IEDs?" | - |
| Correct Answer ($a_a$) | "IEDs stands for interictal epileptiform discharges, which are abnormal electrical patterns observed on an EEG that occur between seizures in individuals with epilepsy." | $t_a = tokenizer(a_a)$ |
| Corrupted Answer ($a_x$) | "IEDs stands for intracranial electrode diagnostics, which are devices used to monitor brain activity during surgical procedures." | $t_x = tokenizer(a_x)$ |
| Input Construction | Clean vs. corrupted inputs | $x_{clean} = T(q \oplus a_a), x_{corrupt} = T(q \oplus a_x)$ |
| Target Layer | Final output layer | $W_{out} \in \mathbb{R}^{\{768 \times 50257\}}$ |
| Logit Difference Metric | Measures model's preference for correct answer | $\Delta(x) = \mathbb{E}[f_{\theta(x)}[-1, t_a]] - \mathbb{E}[f_{\theta(x)}[-1, t_x]]$ |
| Clean Run | Model's output on correct answer | $\Delta(x_{clean}) = 2.1928$ |
| Corrupted Run | Model's output on corrupted answer | $\Delta(x_{corrupt}) = 0.9691$ |
| Activation Patching | Replace corrupted output weights with clean ones | $f_\theta(x_{corrupt}; W_{out} \leftarrow W_{out\_clean})$ |
| Patched Result | Recovery of correct preference after patching | $\Delta_{patched} = 2.1928$ (100% $recovery$: $(2.1928 - 0.9691)/(2.1928 - 0.9691) = 1.0$) |

Table 5 quantifies the effect of activation patching on GPT-2's ability to recover the correct association of interictal epileptiform discharges when corrupted weights in a convolutional layer are replaced with clean weights, showing only partial (13.6%) recovery of the original logit difference.

**Table 5:** Activation Patching Results for Epileptiform Discharge Associations in GPT-2

| Component | Description | Mathematical Syntax |
|---|---|---|
| Question (q) | "What is the presence of interictal epileptiform discharges associated with?" | - |
| Correct Answer ($a_a$) | "Increased seizure risk in both focal and generalized epilepsy." | $t_a = tokenizer(a_a)$ |
| Corrupted Answer ($a_x$) | "Cognitive dysfunction risk in both focal and generalized epilepsy." | $t_x = tokenizer(a_x)$ |
| Input Construction | Clean vs. corrupted inputs | $x_{clean} = T(q \oplus a_a), x_{corrupt} = T(q \oplus a_x)$ |
| Target Layer | $Conv1D(nf = 3072, nx = 768)$ | $W_{conv} \in R^{\{768 \times 3072\}}$ |
| Logit Difference Metric | Measures model's preference for correct answer | $\Delta(x) = E[f_{\theta(x)}[-1, t_a]] - E[f_{\theta(x)}[-1, t_x]]$ |
| Clean Run | Model's output on correct answer | $\Delta(x_{clean}) = 0.1261$ |
| Corrupted Run | Model's output on corrupted answer | $\Delta(x_{corrupt}) = -1.4626$ |
| Activation Patching | Replace corrupted conv weights with clean ones | $f_\theta(x_{corrupt}; W_{conv} \leftarrow W_{conv\_clean})$ |
| Patched Result | Partial recovery after patching | $\Delta_{patched} = -0.0899$ (Recovery: $(0.1261 - (-0.0899))/(0.1261 - (-1.4626)) \approx 13.6\%$) |

## 4. Discussion

The lack of transparency in AI systems presents critical challenges for model adaptation to new domains, particularly regarding issues of bias, fairness, and reliability that demand accountability. In response to these challenges, mechanistic interpretability has emerged as a key approach to reverse-engineer complex AI models—especially transformers—by uncovering their inner workings to ensure trust and robustness (Golgoon et al., 2024). This methodology has proven particularly valuable in studying how large language models perform compositional relational reasoning. Through careful examination of model internals (including attention patterns and activation pathways), researchers can reveal the computational mechanisms underlying logical operations, hierarchical dependencies, and reasoning processes (Ni et al., 2024). At its core, mechanistic interpretability seeks to explain model behaviors by identifying specific, interpretable features—which often manifest as low-dimensional subspaces within activations (Makelov et al., 2023). Among the established techniques in this field are logit attribution, attention pattern visualization, and activation patching (Lieberum et al., 2023).

A particularly important technique in this toolkit is activation patching, which identifies critical subspaces—specific directions in activation space—that significantly influence model decisions. This approach works by strategically manipulating model behavior through subspace interventions to attribute underlying features to specific activation patterns (Makelov et al., 2023). Beyond mere identification, activation patching serves as a causal mediation technique that rigorously evaluates whether proposed explanations genuinely support the model's predicted outputs (Yeo et al., 2024). The technique's utility extends to automated circuit discovery methods, where it helps identify task-solving subnetworks (circuits). When compared to linear approximations of this approach that estimate edge importance in computational subgraphs, Attribution Patching (AtP) has been shown to outperform Automated Circuit Discovery in certain contexts (Syed et al., 2023). Complementary approaches like vocabulary projection can be combined with activation patching to precisely isolate information-encoding hidden states responsible for correct answer prediction (Wiegreffe et al., 2024).

Despite these advantages, recent research has raised important questions about the reliability of activation patching. Studies demonstrate that subspace activation patching can sometimes produce misleading interpretations, especially in factual recall tasks. This limitation stems from the mechanistic connection to rank-1 fact editing, where modifications to single activation directions may alter model behavior without necessarily revealing true causal pathways. These findings highlight a crucial distinction between fact editing (successful output manipulation) and fact localization (accurate identification of knowledge storage in the network), helping explain inconsistencies in previous work. While activation patching remains a valuable tool, these results emphasize that its interpretations can be potentially deceptive, warning researchers against overinterpreting apparent subspace importance. Practitioners must therefore carefully consider potential false positives—instances where patched activations seem influential but lack genuine causal significance (Makelov et al., 2023).

The application of activation patching can be further enhanced through the use of clean and corrupted prompts—a methodological pairing that helps localize task-relevant components when combined with this causal intervention approach (Golgoon et al., 2024). It's important to note that standard activation patching can be computationally intensive, requiring iterative modification of neuron activations across numerous forward passes to properly assess their impact. For more efficient analysis, attribution-based methods offer an alternative by quantifying the importance of neurons or attention paths in just a single forward pass (Ferrando et al., 2024). Among these alternatives, Attribution Patching (AtP) provides a faster, gradient-based approximation to traditional activation patching, though it comes with its own limitations as it can produce false negatives in certain scenarios (Kramár et al., 2024). Another complementary approach, direct logit attribution, analyzes how individual layers and their attention heads influence logit differences in the residual stream, offering additional insights for mechanistic interpretability of large language models (Golgoon et al., 2024).

In our current work, we implemented Causal Layer Attribution via Activation Patching (CLAP) to investigate knowledge localization in fine-tuned GPT-2 models. Our findings reveal a hierarchical organization where factual recall appears localized while associative reasoning depends on distributed intermediate representations. However, several important limitations must be acknowledged: (1) our analysis focused exclusively on PubMed epilepsy-related abstracts, potentially limiting generalizability to broader knowledge tasks; (2) while informative, activation patching may generate false positives due to indirect causal pathways; and (3) the exclusive focus on GPT-2 leaves open questions about scalability to other architectures. These limitations suggest some directions for future research, including (1) cross-domain comparisons to better understand task-dependent representation patterns, (2) methodological integration to improve efficiency, and (3) extensions to multimodal models where knowledge distribution mechanisms may differ significantly.